\documentclass[a4paper,fleqn]{cas-dc}

\usepackage[numbers]{natbib}
\usepackage{pifont}
\usepackage{subcaption}
\usepackage{amsmath}

\ExplSyntaxOn
\cs_gset:Npn \__first_footerline:
  { \group_begin: \small \sffamily \__short_authors: \group_end: }
\ExplSyntaxOff 

\def\tsc#1{\csdef{#1}{\textsc{\lowercase{#1}}\xspace}}
\tsc{WGM}
\tsc{QE}
\tsc{EP}
\tsc{PMS}
\tsc{BEC}
\tsc{DE}

\begin{document}
\let\WriteBookmarks\relax
\def\floatpagepagefraction{1}
\def\textpagefraction{.001}

\shorttitle{Computer Vision-based Characterization of Large-scale Jet Flames}

\shortauthors{C. Pérez-Guerrero et~al.}

\title [mode = title]{Computer Vision-based Characterization of Large-scale Jet Flames using a Synthetic Infrared Image Generation Approach}        


\author[addr1]{Carmina Pérez-Guerrero} \corref{contrib}

\author[addr1]{Jorge Francisco Ciprián-Sánchez} \corref{contrib}

\author[addr2]{Adriana Palacios}[orcid=0000-0002-6972-5988]\cormark[1]

\cortext[cor]{C. author(s):  adriana.palacios@udlap.mx, gilberto.ochoa@tec.mx}

\author[addr1]{Gilberto Ochoa-Ruiz} [orcid=0000-0002-9896-8727] \cormark[1] 

\author[addr1]{Miguel Gonzalez-Mendoza}

\author[addr3]{Vahid Foroughi}

\author[addr3]{Elsa Pastor}

\author[addr4]{Gerardo Rodriguez-Hernandez}

\address[addr1]{Tecnologico de Monterrey, School of Engineering and Sciences, Jalisco, 45138, Mexico}

\address[addr2]{Universidad de las Américas Puebla, Department of Chemical, Food and Environmental Engineering, Puebla, 72810, Mexico}

\address[addr3]{Universitat Politècnica de Catalunya. EEBE, Eduard Maristany 16, 08019 Barcelona. Catalonia, Spain}

\address[addr4]{CIATEQ A.C., Artificial Intelligence Laboratory, Jalisco, 45131, México.}

\cortext[contrib]{Authors contributed equally.}

\begin{abstract}
Among the different kinds of fire accidents that can occur during industrial activities that involve hazardous materials, jet fires are one of the lesser-known types. This is because they are often involved in a process that generates a sequence of other accidents of greater magnitude, known as domino effect. Flame impingement usually causes domino effects, and jet fires present specific features that can significantly increase the probability of this happening. These features become relevant from a risk analysis perspective, making their proper characterization a crucial task. Deep Learning approaches have become extensively used for tasks such as jet fire characterization; however, these methods are heavily dependent on the amount of data and the quality of the labels. Data acquisition of jet fires involve expensive experiments, especially so if infrared imagery is used. Therefore, this paper proposes the use of Generative Adversarial Networks to produce plausible infrared images from visible ones, making experiments less expensive and allowing for other potential applications. The results suggest that it is possible to realistically replicate the results for experiments carried out using both visible and infrared cameras. The obtained results are compared with some previous experiments, and it is shown that similar results were obtained.

\end{abstract}



\maketitle

\section{Introduction}
\label{sect:intro}
The production, handling, and transport of hazardous materials in the chemical industry have increased during the last decades. This has led to an increase in the frequency with which serious major accidents occur. The consequences that these have had on people and the environment have highlighted the need to improve industrial safety measures. This has been reflected in the implementation of new legislation such as the Seveso Guideline~\cite{OJEC1982}~\cite{OJEC1996}~\cite{OJEC2003} on prevention of major accidents in certain industrial activities. Thus, the development of research studies on the main characteristics, mathematical modeling, and estimation of consequences of these major accidents, occurring in the chemical industry around the world is of paramount importance.

Several studies have shown that, after containment losses, one of the most common major accidents is fire. Among the various types of major fires that can occur, jet fires are one of the lesser-known types until just recently. Jet fires are a particular type of serious major fire accident that can occur both in fixed installations (process plants, storage, or pipe networks) and, less frequently, during transport (trucks or tank cars). Although jet fires occur with some frequency, many of them are not reported, since they are often accompanied by other events of greater magnitude, such as explosions. It is important to note that jet fires are also capable of generating a sequence of other accidents, called a domino effect; since their duration and the heat emitted are high enough to cause it ~\cite{Casal18}. An example of an incident caused by a jet fire is the accident that occurred in 2007 at the Valero McKee Refinery in Sunray, Texas ~\cite{Valero07}. A leak in the propane deasphalting unit caused a domino effect that spread quickly, following the collapse of a major pipe rack that was carrying flammable hydrocarbons. The accident caused four injured, three of them with serious burns, and the extended shutdown of the refinery.


Jet fires can be evaluated based on specific features that significantly increase the probability of flame impingement, which in turn can lead to lethal domino effects. These features become relevant from a risk analysis perspective, which magnifies the value of their proper characterization. A great majority of consequence analyses regarding jet fires, that evaluate both geometric features and radiation loads, employ physically-based models, such as Computational Fluid Dynamics (CFD)~\cite{colella2020}, or empirical models, which are based on empirical correlations developed from a set of experiments. However, most empirical models use a simplified flame geometry and fixed emissive power to determine the thermal radiation load on a target. Furthermore, Lattimer et al.~\cite{Lattimer20} presented a general overview of machine learning techniques for low-cost and high fidelity fire predictions where it was found that machine learning has the capacity to provide full-field predictions faster than CFD simulations by 2 to 3 orders of magnitude.

Deep Learning, a subset of Machine Learning, has become extensively used and increasingly influential when addressing complex tasks. Convolutional Neural Networks (CNNs), for example, have shown notable performance in complex computer vision tasks, such as image recognition, object detection, and semantic segmentation ~\cite{Litjens_2017}. A recent stride in the use of CNNs for the characterization of jet fires is explored in Pérez-Guerrero et al. ~\cite{Perez-Guerrero22}, where geometric features of the flames are extracted from segmentation masks of their radiation zones. The models explored, however, are heavily dependent on the amount of data and the quality of the labels. Data acquisition of jet fires is done at a high cost since it involves expensive experiments where both visible and infrared (IR) images of the flames need to be captured, and then the collection of the jet flame images needs to be paired and jointly analyzed.

In recent times, Generative Adversarial Networks (GANs) have displayed state-of-the-art performance for the task of synthetic image generation. There are examples of the use of this capability for data augmentation tasks in different domains, such as medical image classification~\cite{Wu18}, speech recognition~\cite{Sheng19}, sensor simulation~\cite{Milz18}, amongst others.

For the domain of synthetic fire image generation, there has been recent work regarding the use of GANs for the task of visible fire image synthesis~\cite{Yang22,Park20,Zhikai19}, wildfire smoke generation~\cite{Mao21}, and artificial wildfire infrared (IR) image generation and fusion~\cite{Ciprian-Sanchez21}. In particular, the work by Ciprián-Sánchez et al.~\cite{Ciprian-Sanchez21} is of special relevance to the present paper as it is, to the best of our knowledge, the first one to employ GANs to generate artificial IR fire images from visible ones. 

In the context of jet flame risk management, the generation of synthetic IR images through visible samples has the potential of substantially diminishing the costs and complexity associated with the development and implementation of jet flame segmentation and characterization systems.

This paper presents, what is to the best of our knowledge, the first use of GANs to generate artificial IR jet flame images through visible ones. Taking as a precedent the work by Ciprián-Sánchez et al.~\cite{Ciprian-Sanchez21}, the Pix2Pix GAN model is implemented. First proposed by Isola et al.~\cite{Isola17}, this model is one of the most widely used for image-to-image translation. This model is trained on a dataset of visible-IR jet flame image pairs and validate its effectiveness for artificial IR jet flame image generation through four different image quality and similarity metrics: (i) image entropy (EN)~\cite{Shannon48}, (ii) correlation coefficient (CC)~\cite{Xu22}, (iii) peak signal-to-noise ratio (PSNR)~\cite{Korhonen12}, and (iv) the structural similarity index measure (SSIM)~\cite{Wang04}.

Next, the Deep Learning (DL)-based jet flame segmentation and characterization system proposed by Pérez-Guerrero et al.~\cite{Perez-Guerrero22} is implemented on both real and artificial IR images. Two different CNNs, namely UNet and Attention UNet, are used to obtain a segmentation mask of the radiation zones within the flames. This segmentation is then used to extract the jet fire's geometrical information and the results are compared against the real experimental data obtained form large-scale jet flames. Finally, an evaluation is done on the errors obtained from artificial and real IR images.

Furthermore, an assessment and evaluation was done regarding the feasibility and advantages of implementing GAN-based approaches to mitigate the costs and complexity associated with IR image-based jet flame risk management systems.

The rest of this paper proceeds as follows. Section~\ref{sect:prev_work} introduces previous works on jet flame characterization systems, the role of computer vision for the improvement of these systems, and introduces relevant works related to GAN-based fire image synthesis. Section~\ref{sect:approach} presents the proposed approach for the image acquisition and preprocessing, and presents the DL pipeline for the artificial IR image generation and jet flame segmentation and characterization. Next, in Section~\ref{sect:experiments_discussion}, the experimental results are presented and discussed. Finally, Section~\ref{sect:conclusions} concludes the paper and outlines potential avenues for further research.

\section{Previous work}
\label{sect:prev_work}

\subsection{Risk assessment and modeling}
\label{subsect:risk_modeling}
The production, handling, and transport of hazardous materials in the chemical industry have increased during the last decades. This has led to an increase in the frequency with which serious major accidents occur. The consequences that these have had on people and the environment have highlighted the need to improve industrial safety.

There are certain industrial activities such as the production, handling, storage, or transportation of hazardous materials, that can be involved in serious major fire accidents that can threaten the human health, have a high environmental impact, and can cause important property damage. Therefore, as part of risk assessment, it is important to identify the optimal space between equipment and structures by evaluating the size and shape of the flames. Furthermore, the thermal radiation of fire incidents can be analyzed to monitor the heat transfer process to nearby objects. Therefore, fast and reliable tools to obtaining the shape and proportions of a jet fire, as well as their radiation properties is extremely important from a risk management point of view~\cite{Kashi2020}. 

Early risk assessment models are based on algebraic expressions that assume an idealized flame shape with uniform surface radiation emissive power. Some examples are the proposal of a cylindrical crone by Croce and Mudan~\cite{croce1986}, or a frustum of a cone by Chamberlain~\cite{Chamberlain87}. However, the lack of consideration to the geometry of the fire system on these methods made them unreliable when dealing with barriers or equipment, and thus several other empirical models have been explored to evaluate the consequences of the industrial hazards that fire accidents represent. 

For instance, different radiation and turbulence models, along with CFD simulations, were explored by Kashi and Bahoosh~\cite{Kashi2020} to evaluate how firewalls or equipment around the flame can affect the radiation distribution and temperature profile of the jet fire. The Delichatsios' model was explored by Guiberti et al.~\cite{Guiberti20} to obtain the flame height for subsonic jet flames at elevated pressure; however, this model by itself predicts well around 20\% of the flame height in these cases. Mashhadimoslem et al.~\cite{mashhadimoslem20} explored two Artificial Neural Networks (ANN) to estimate the jet flame lengths and widths based on the mass flow rates and the nozzle diameters. The two methods presented negligible discrepancies between them and can be used in place of CFD methods, which in turn require more computational time and resources. Finally, Wang et al.~\cite{WANG21} implemented an Otsu method to determine the flame's geometric features and to estimate the flame extension area based on the flame intermittency probability. The flame pattern and color are observed to evolve according to an increase in exit velocity and a decrease in exit-plate spacing.

Despite the great performance of some of these methods, they heavily depend on domain expertise. Furthermore, since fire has very dynamic geometry and shapes, some models grow in uncertainty when dealing with different conditions. An automatic monitoring system that uses computer vision to analyze the actual shape and geometry of the flame could add significant information that could improve e the results of the mentioned models.

\subsection{The role of computer vision in fire characterization}
\label{subsect:cv_role_fire_charact}
There has been previous work that explore the use of traditional computer vision methods to perform fire characterization for risk assessments. To mention some examples, Janssen and Sepasian~\cite{Janssen18} presented an automatic flare-stack-monitoring system that employs computer vision to separate the flare from the background through temperature thresholding. To enhance the observed temperature profile of the flare, false colors are added, representing temperature regions within the flare. Progressing on that, Zhu et al.~\cite{zhuli20} proposed an infrared image-based feedback control system that uses an improved maximum entropy threshold segmentation method to identify and localize the fire for automatic aiming fire extinguishing and continuous fire tracking. Alternatively, Gu et al.~\cite{Gu20} presented a Computer Vision-based Monitor of Flare Soot, abbreviated as VMFS, that is mainly composed of three stages that sequentially apply a tuned color channel to localize the flame, a saliency detection using K-means to determine the flame area, and finally, use the center of the flame to search for a potential soot region.

These different traditional computer vision approaches tend to show an excellent performance when dealing with cases similar to the data that they were developed on; however, it is necessary to choose which features are important for each specific problem, so various aspects involve domain knowledge and long processes of trial and error. Deep Learning architectures, in contrast, are able to discover the underlying patterns of the images and automatically detect the most descriptive and salient features with respect to the problem. This is further proven in the work presented by Pérez-Guerrero et al.~\cite{Perez-Guerrero21}, where a comparison was done between traditional computer vision algorithms and Deep Learning models to evaluate a segmentation approach for the characterization of fire, based on radiation zones within the jet flames.

In spite of that, Deep Learning architectures are reliant on the size and quality of the training data. Depending on the task domain, data acquisition can become difficult and expensive, leading to limited training data sets that can cause over-fitting and negatively affect the model's generalization capacity. Since Deep Learning architectures contain a high number of parameters with intricate inter relationships, it is considered quite complex to manually tweak these models, making them a black box in comparison to the transparency of the traditional computer vision approaches~\cite{Mahony19}. However this issue can be tackled through the use of several data augmentation strategies, that increase the training examples available.

\subsection{GANs for synthesizing IR imagery for  image analysis}
\label{subsect:GANS_artificial_imgs}
The use of GANs for fire imagery generation is a nascent field, with works such as the ones proposed by Zhikai et al.~\cite{Zhikai19}, Park et al.~\cite{Park20}, and Zhikai et al.~\cite{Yang22} focusing on the generation of synthetic visible fire images to augment datasets used mainly for the training of fire detection or segmentation methods.

Similarly, the work by Mao et al.~\cite{Mao21} focuses on using GANs to improve synthetic wildfire smoke images to a photorealistic level. The authors employ their proposed approach to perform data augmentation for early wildfire smoke detection systems.

The work by Ciprián-Sánchez et al.~\cite{Ciprian-Sanchez21} focuses on visible-infrared wildfire image fusion. The FIRe-GAN model proposed by the authors first generates artificial IR wildfire images through visible ones, fusing the real visible and the artificial IR ones to produce a final fused image. This work introduces the first example of GANs for image-to-image translation of fire images (in this case, using visible images to generate approximate IR ones).

The ability to produce approximate IR images through visible ones is very desirable for jet flame risk management. It has the potential to reduce the costs and complexity of developing and deploying IR-based jet flame segmentation and characterization systems, as visible cameras are generally cheaper, more flexible, and widely available.

For the present work, and taking as a precedent the FIRe-GAN model~\cite{Ciprian-Sanchez21}, the Pix2Pix model is implemented, one of the most widely used image translation architectures, first proposed by Isola et al.~\cite{Isola17} to perform visible-to-infrared image translation of jet flame imagery. 

\subsection{Justification and outline of the contribution}
\label{subsect:justification_contribution}
Jet fires, compared to other major fire accidents, have a reduced damage radius; however, this does not mean that their effects cannot be catastrophic. Since they originate very high heat fluxes, in case flames reach other equipment, they can cause a domino effect, triggering a sequence of more severe accidents. For this reason, it is of great interest to study their dimensions to establish their effects and consequences to eventually propose the required safety distances and zones.

The properties of jet fires depend on the composition of the fuel, their release condition (i.e., gas, liquid or two-phase flow), the fuel release speed and certain atmospheric conditions, such as the direction and speed of the wind, among other variables. To study jet fires properly, it is necessary to know the terms stated previously, while also taking into account their geometrical features. Some relevant geometric characteristics are the distance between the fuel release point (i.e., nozzle) and the visible tip of the flame, and the total flame area. These features are necessary to determine the probability of thermal radiation, which can be very high at short distances, and flame impingement on nearby objects.

The creation and evaluation of models for the extraction of the mentioned jet fire properties would require a considerable amount of data from jet fire experiments. The composition of the fuel, the release condition and release speed can be established for these experiments and the atmospheric conditions can be monitored during them; however the dynamic nature of the resultant fire shape can be only captured and evaluated in that moment. Visual cameras with formats such as Video Home System (VHS) are the most common method of fire video acquisition, however visible light can be altered by the atmospheric conditions or blocked by some consequences of fire, such as smoke. A more reliant evaluation of the flame shape would then come from Infrared (IR) images, however IR cameras are more expensive than visible ones, therefore  IR video surveilance is not always available. Furthermore, additional work would be needed to pair both visible and IR images for their joint analysis. All of this adds to the already high cost and laboriousness of fire experimentation, which is why the research presented in this paper explores a solution for the generation of IR images from visible ones. These results could aid in significantly speeding up the experimentation process and eliminating the dependency on IR camera availability, contributing to future research regarding different fire incidents.

\section{Proposed approach}
\label{sect:approach}

\subsection{Image acquisition}
\label{subsect:image_acq}

\begin{figure}[!htbp]
  \centering
  \includegraphics[width=9.5cm]{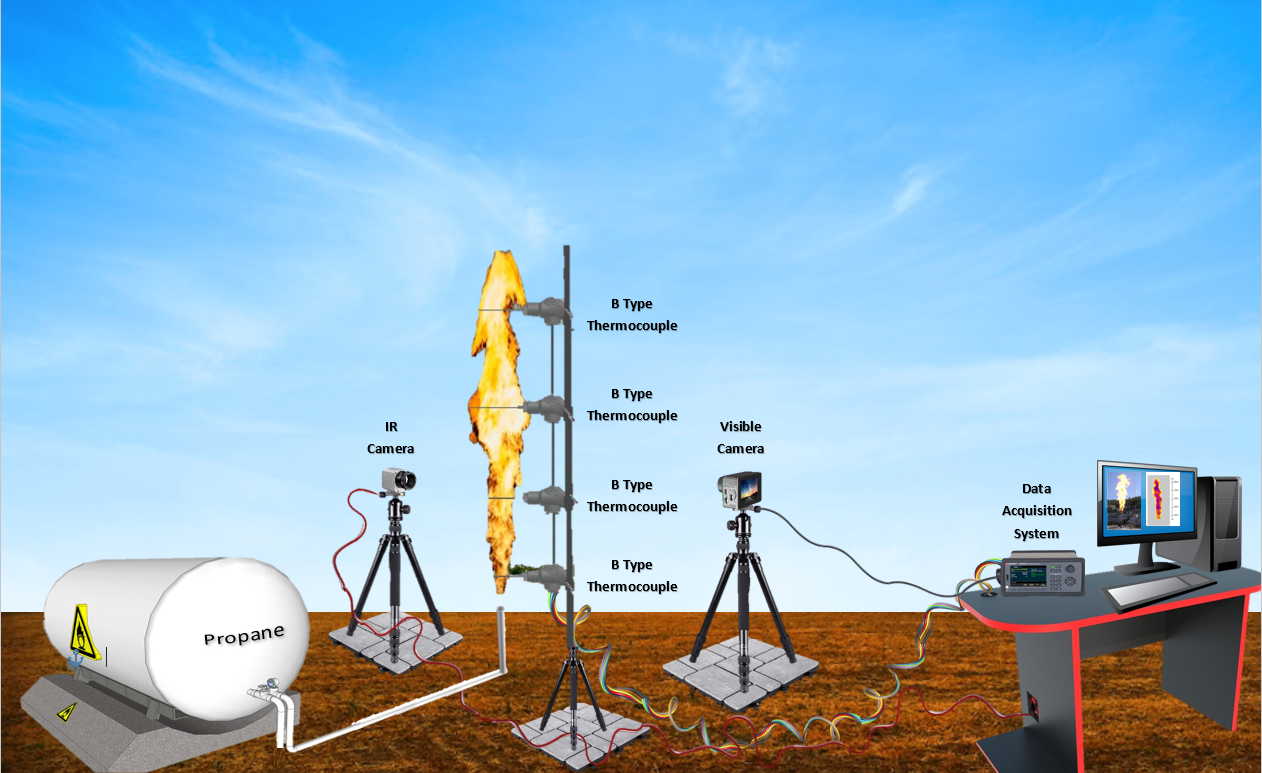}
  \caption{A schematic view of the experimental set-up for the large-scale jet fires analysed in the present study.}
  \label{fig:setup}
\end{figure}

Vertical large-scale jet fires, released in the open-field, have been analysed in the present study. The jet fire experiments involve subsonic and sonic jet flames of propane, discharged through nozzles with a diameter of 12.75 mm, 15 mm, and 43.1 mm. The mass flow rates ranged between 0.03 kg/s to 0.28 kg/s. The experiments were filmed with two video cameras registering visible light (VHS) and an infrared thermographic camera (Flir Systems, AGEMA 570). Twenty-five digital images per second were obtained from the visible spectrum, using two cameras with resolutions of 384 x 288 pixels and 320 x 240 pixels, respectively; while four images per second were obtained from the infrared (IR) camera with a resolution of 640 x 480 pixels. The two visible cameras were located orthogonally to the flame, and one of them was located next to the infrared thermographic camera. The average wind speed was 0.4 m/s and the average ambient temperature was 28°C for all the experimental tests performed. The experimental set-up is shown by the schema in Fig.~\ref{fig:setup} Further details of the experimental set-up can be found in ~\cite{Adriana11}.

\subsection{Dataset}
\label{subsect:dataset}
The training dataset used for the Pix2Pix model came from two different experiments. The "JFP 05 11" experimental flames were discharged through a nozzle of 12.75 mm and resulted in 136 pairs of visible and IR images. The "JFP 05 12" experimental flames were discharged through a nozzle of 43.1 mm and resulted in 59 pairs of visible and IR images. Both experiments amounted to 195 paired images were the visible images had a resolution of 384 x 288 pixels, while the IR images had a resolution of 640 x 480 pixels. It was important that the flames were clearly visible during the experiments.

The dataset used for validation also came from two different experiments. Both experimental flames were discharged through a nozzle of 15 mm and the main differences between them were the temperature, the humidity, and the mass flow rate range. These differences are summarized in Table \ref{tab:15mmexp}. The experimental data of the first test covers up to 52 images; while the experimental data for the second test covers up to 37 images, which resulted in a total of 89 images obtained from visible light video with a resolution of 320 x 240 pixels.

\begin{table}[!htbp]
\centering
\caption{Flame characteristics of the experiments used for the validation dataset.}
\label{tab:15mmexp}
\begin{tabular}{ccccc}
\hline
\textbf{Experiment} & \textbf{\begin{tabular}[c]{@{}c@{}}Ambient\\ Temperature\\ (K)\end{tabular}} & \textbf{\begin{tabular}[c]{@{}c@{}}Relative\\ Humidity \\ (\%)\end{tabular}} & \textbf{\begin{tabular}[c]{@{}c@{}}Mass \\ Flow Rate\\ (kg/s)\end{tabular}} & \textbf{\begin{tabular}[c]{@{}c@{}}Total \\ Images\end{tabular}} \\ \hline
JFP 05 11 & 301.15 & 55  & 0.14 - 0.22 & 52 \\ \hline
JFP 05 12 & 298.15 & 50 & 0.24 - 0.28 & 37 \\ \hline
\end{tabular}
\end{table}

\subsection{Image preprocessing}
\label{subsect:img_preprocessing}
\begin{figure}[!htbp]
  \centering
  \includegraphics[width=6cm]{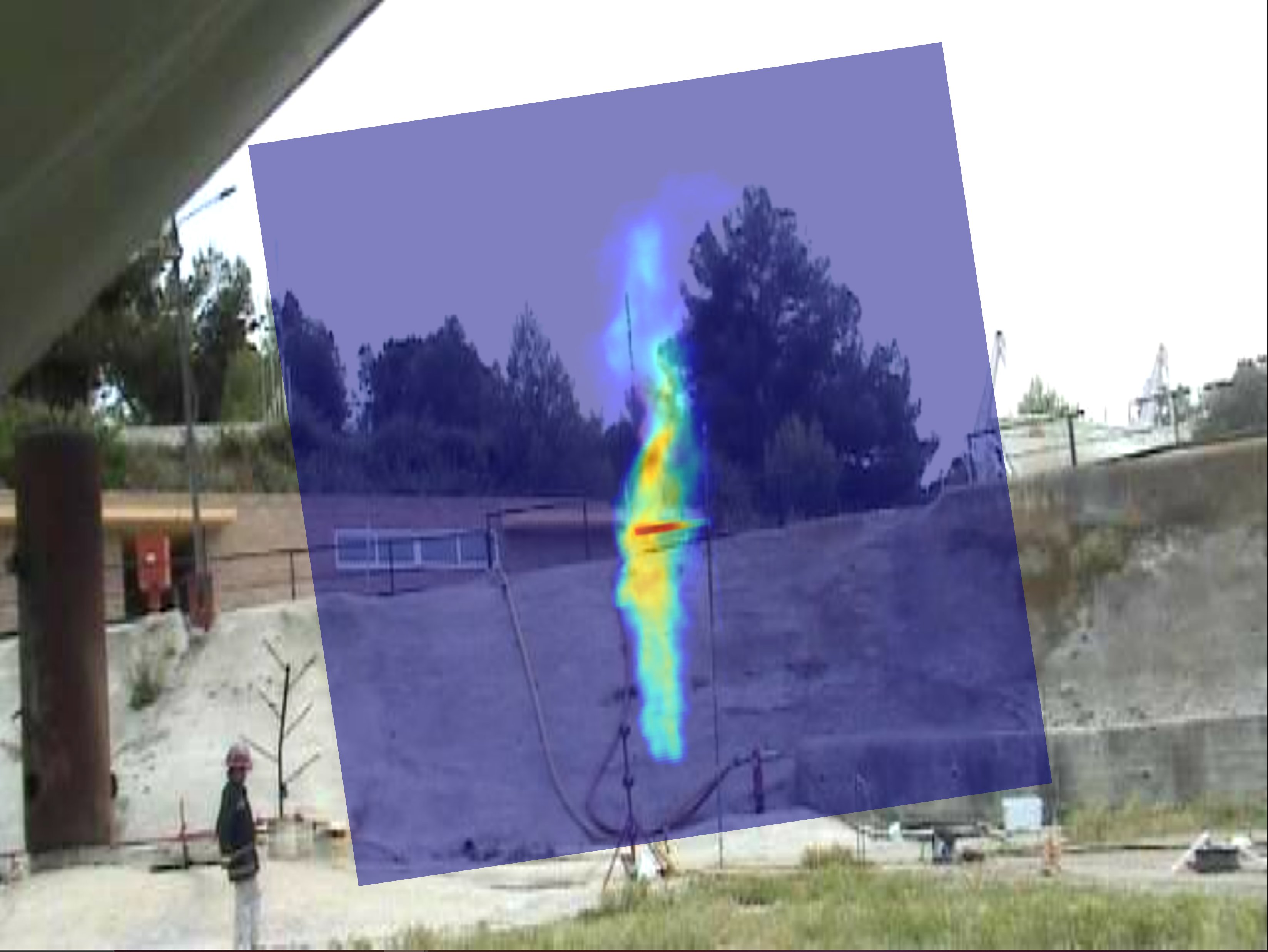}
  \caption{A visualization of the framing difference between a visible VHS frame and a thermal IR frame of the same jet fire.}
  \label{fig:framing-diff}
\end{figure}

The frames from both the VHS visible light video and the thermal IR video of the jet fire experiments needed to be paired for the experiments performed in the present work. The VHS video was saved in a AVI format, while the infrared data was saved as mat files that contain the temperature values in a matrix, using an emissivity value of 0.3 \cite{Palacios12}. The mat files were exported as PNG images and turned into a video sequence with AVI format through a video editing software. With the two videos in the same format, the focus was then the different framing between the two videos, which can be observed in Fig.~\ref{fig:framing-diff}. This was fixed through the use of a video editing software and using the thermocouples and pipe nozzle as reference for the necessary transformations based on two arbitrary frames. Finally, the last problem to deal with was the significant difference of frame rate between the two video. The VHS video has roughly 30 frames per second, while the IR video has roughly 9 frames per second.

One approach was to track them one by one through a video editing software. However that process is very time consuming, so a new approach was employed to recreate the higher frame rate video as a matching lower frame rate video. This was done by matching the timestamps of the two videos, using a frame and timestamp form the IR video as reference and comparing it to frames and timestamps from the VHS video. However, this approach is only feasible when both videos start and end at the same time, while also having a consistent frame rate, which was the case for the present data set.

\subsubsection{Artificial IR image generation}
\label{subsubsect:pix2pix_preprocessing}

The images were resized to a height of 1024 and a width of 512 pixels to ensure that the input and output dimensions of the images fed the Pix2Pix model remain the same. Zero padding is added to the images until the desired size is reached; this allows to keep the original image dimensions for the segmentation and characterization steps, in which said padding is removed.

\begin{figure}[!htbp]
  \centering
  \includegraphics[scale=0.38]{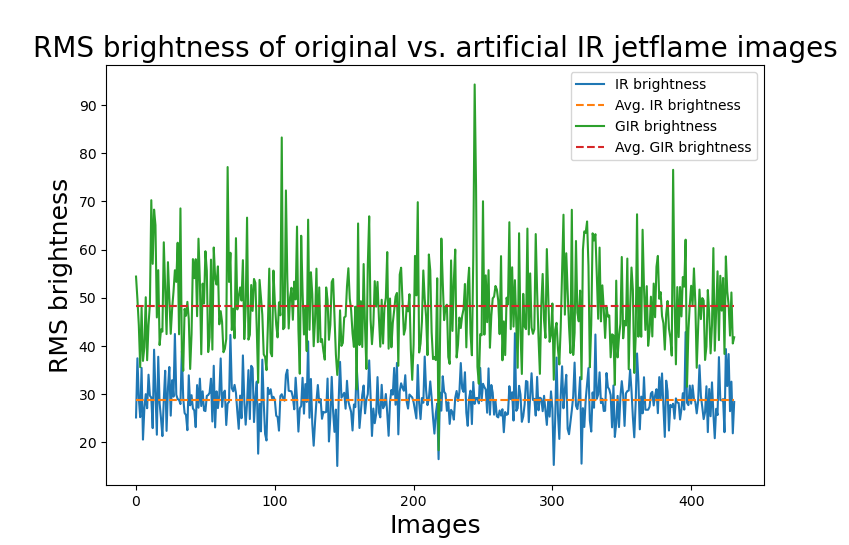}
  \caption{RMS brightness analysis and comparison between the real and synthetic IR images.}
  \label{fig:brightness_analysis}
\end{figure}

Furthermore, as it is presented in more detail in Section~\ref{sect:experiments_discussion}, the Pix2Pix model occasionally generated synthetic IR images that displayed a considerably higher brightness level when compared to the real ones. So, the Root Mean Square (RMS) brightness level of the real and artificial IR images was analyzed, noticing a clear pattern of brighter artificial images. Figure~\ref{fig:brightness_analysis} shows the average and per-image RMS brightness values of the real and synthetic IR images.

Thus, the average brightness level was calculated for both types of images and established the difference between both averages (for this case, $20$) as a threshold for a post-processing step to adapt the brightness level of the generated images. The formula that was defined to adapt the brightness of the generated IR images is as follows:

\begin{equation}
\label{eq:brightness_adjustment}
B(I_a) = 
\begin{cases} 
\frac{1.5 B(I_r)}{B(I_a)} , &\textrm{if $B(I_a)-B(I_r) > 20$}\\
B(I_a), &\textrm{otherwise,}
\end{cases}
\end{equation}

where $B(I_a)$ represents the RMS brightness of the artificial image and $B(I_r)$ the RMS brightness of the real image.

\subsubsection{Jet flame segmentation and characterization}
\label{subsubsect:seg_charact_preprocessing}

The segmentation masks of jet flames are obtained from the implementation of UNet and Attention UNet architectures for the problem of jet fire radiation zone segmentation described in Pérez-Guerrero et al.~\cite{Perez-Guerrero21}. These architectures were found to be the best performing ones among eight different methods and models explored. The original IR images and the generated IR images are both used as input for each of the models to obtain the segmentation masks of each set of images.

The jet flame area and the distance between the fuel release point and the visible tip of the flame, described as the total height, are extracted from the segmentation masks following the procedure presented in Pérez-Guerrero et al.~\cite{Perez-Guerrero22}, which is based on the segmentation mask contour, a reference point to the outlet of the flame, and a ratio of the pixels per metric of the image.

\subsection{Deep Learning methods and pipeline description}
\label{subsect:DL_pipeline}
\begin{figure*}[!htbp]
  \centering
  \includegraphics[scale=0.7]{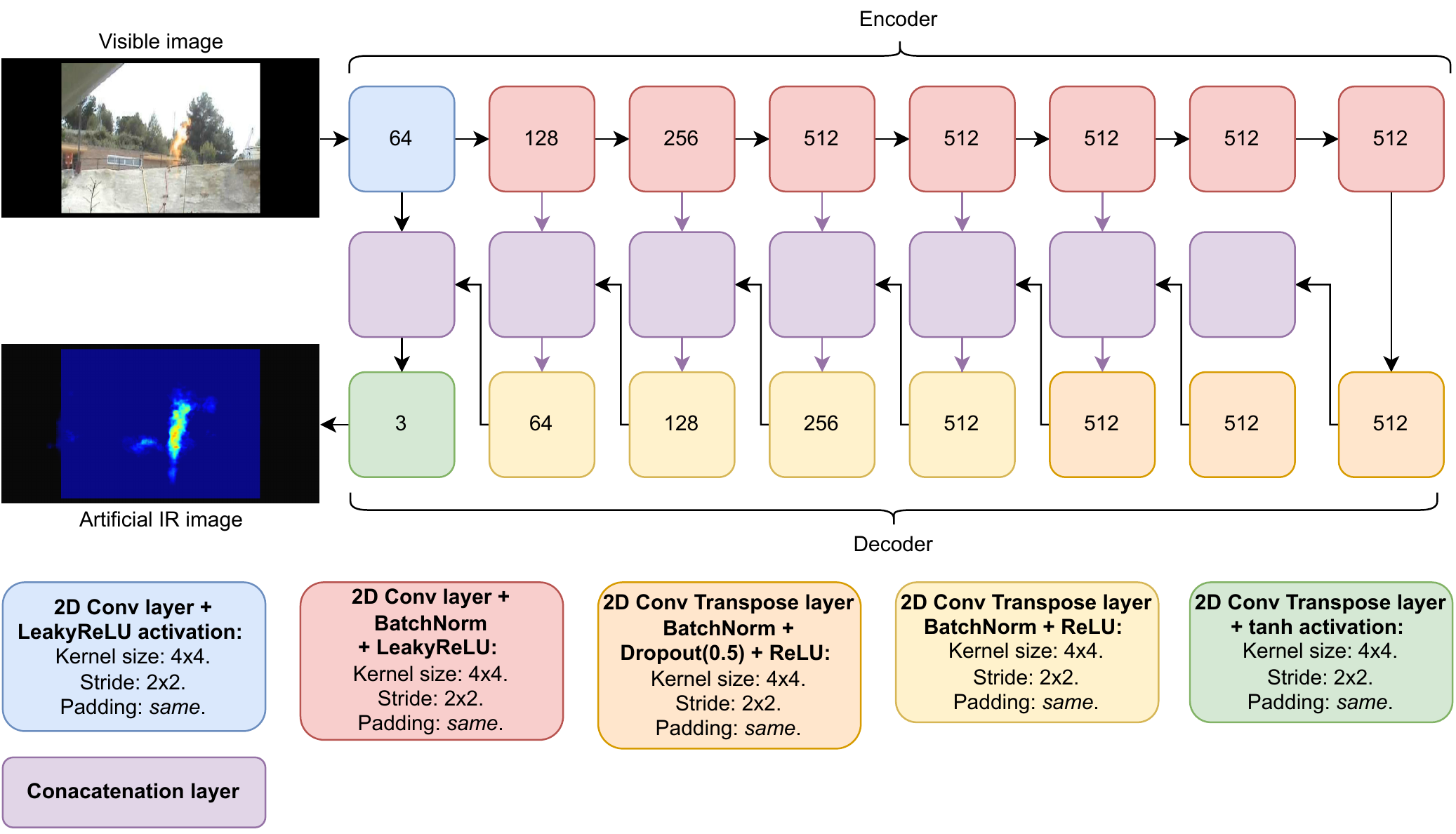}
  \caption{The implementation of the generator of the Pix2Pix model by Isola et al.~\cite{Isola17} for visible-to-infrared jet flame image translation. The numbers inside the squares represent the number of filters per layer. The color-coded squares bellow the main figure detail the convolution module structure of the layers.}
  \label{fig:pix2pix_gen_schema}
\end{figure*}

\begin{figure*}[!htbp]
  \centering
  \includegraphics[scale=0.7]{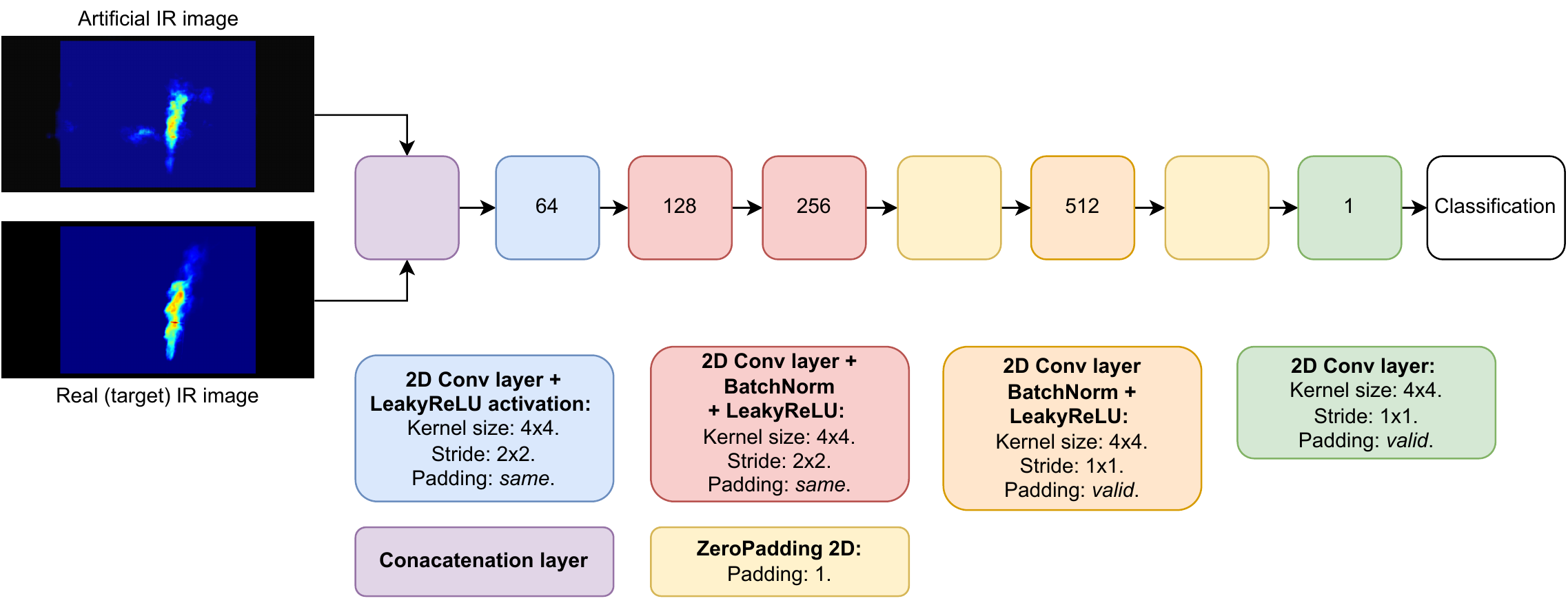}
  \caption{The implementation of the discriminator of the Pix2Pix model by Isola et al.~\cite{Isola17} for visible-to-infrared jet flame image translation. The numbers inside the squares represent the number of filters per layer. The color-coded squares bellow the main figure detail the convolution module structure of the layers.}
  \label{fig:pix2pix_disc_schema}
\end{figure*}

The first step in the pipeline of the proposed approach is the generation of the artificial IR jet flame images. To generate the artificial IR jet flame images, the visible jet flame images were pre-processed and used as input for the Pix2Pix model. Figure~\ref{fig:pix2pix_gen_schema} shows the in-detail architecture of the implementation of the Pix2Pix generator, and Figure~\ref{fig:pix2pix_disc_schema} of the Pix2Pix discriminator.

This model generates the artificial IR images, which are then cropped and post-processed as required by the segmentation and characterization models. In Section~\ref{subsubsect:pix2pix_preprocessing} the pre-and post-processing steps for this stage are described.

The second step in the pipeline of the proposed approach is the semantic segmentation of both, the original and the artificial IR jet flame images. To generate the segmentation masks of jet flames, the IR images are used as input for the UNet and Attention UNet segmentation models, which were previously trained on a dataset of horizontal jet fires, for the segmentation of radiation zones within the flames, as described in Pérez-Guerrero et al. ~\cite{Perez-Guerrero21}. 

\begin{figure*}[!htbp]
  \centering
  \includegraphics[scale=0.55]{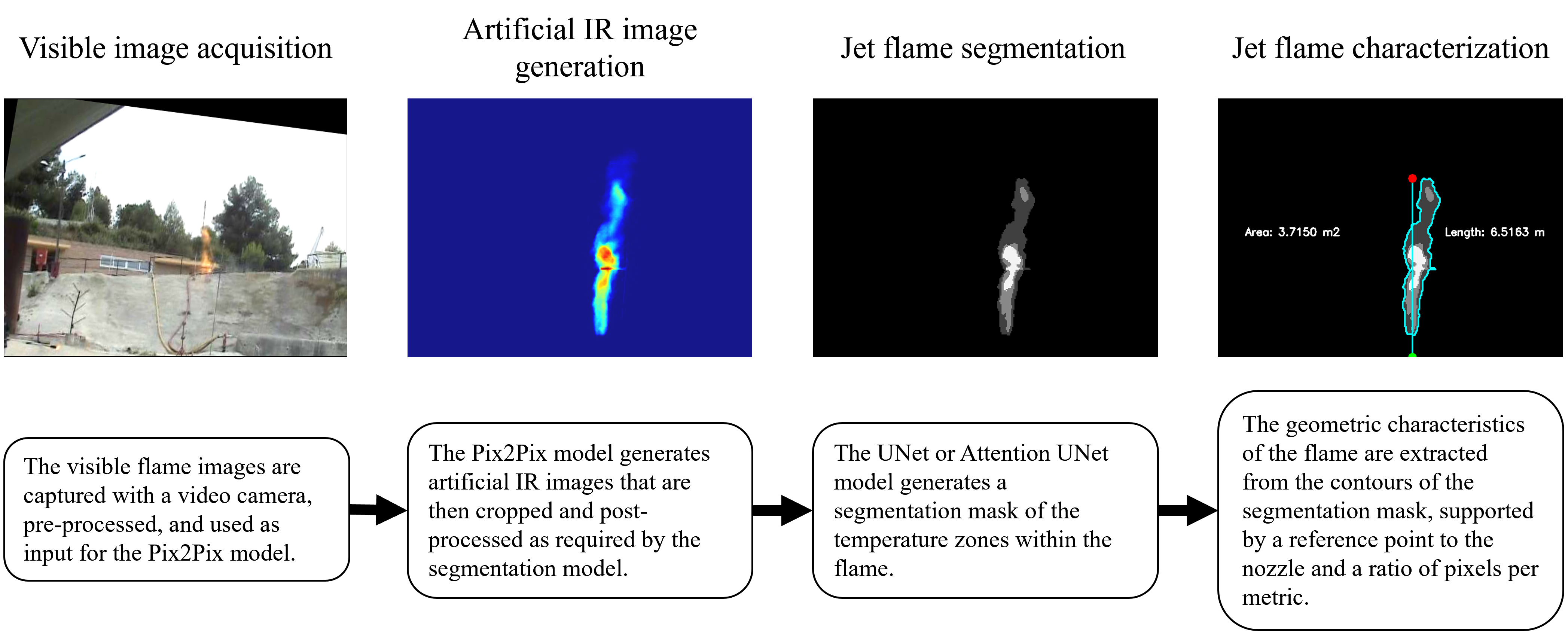}
  \caption{A visual example of the process done through the pipeline of the proposed approach.}
  \label{fig:pipeline}
\end{figure*}

The third and final step in the pipeline of the proposed approach is the extraction of fire geometric characteristics from the obtained segmentation masks. This process is similar to the one described in Pérez-Guerrero et al.~\cite{Perez-Guerrero22}, where the contour of the segmentation mask and a reference point to the nozzle, that serves as outlet of the flame, are used to obtain the jet flame area and the total height. An overall visualization of this process can be observed in Fig. \ref{fig:pipeline}.

\subsection{Evaluation metrics}
\label{subsect:eval_metrics}

\subsubsection{Artificial IR image generation}
\label{subsubsect:artificial_ir_generation}
For the present paper, four widely used image similarity and quality metrics for the image generation task~\cite{Cheon21,Xu22,Roberts08} were employed: image entropy (EN)~\cite{Shannon48}, correlation coefficient (CC)~\cite{Xu22}, peak signal-to-noise ratio (PSNR)~\cite{Korhonen12}, and the structural similarity index measure (SSIM)~\cite{Wang04}.

The EN uses Shannon's formula~\cite{Shannon48} to quantify the amount of information per pixel~\cite{Roberts08}. It is defined as follows: 

\begin{equation}
\label{eq:information_entropy}
EN = - \sum_{l=1}^{L-1} p_l \log_2 p_l,
\end{equation}

where $L$ stands for the gray levels of the image, and $p_l$ represents the proportion of gray-valued pixels $i$ in the total number of pixels~\cite{Zhao20_GAN}. The larger the EN, the more information is in the generated image.

The CC measures the degree of linear correlation between the generated image and a source reference one. It is defined as follows: 

\begin{equation}
\label{eq:correlation_coefficient}
CC(X,Y) = \frac{Cov(X,Y)}{\sqrt{Var(X) Var(Y)}},
\end{equation}

where $Cov(X,Y)$ is the covariance between the generated image and the reference one, and $Var(X)$, $Var(Y)$ represent the variance of the two images~\cite{Zhao20_GAN}. The larger the value of CC, the higher the correlation between the generated and the reference images. 

The possible values for the CC are in the range of $[-1,1]$. A value of $-1$ means that the images are inversely correlated, a value of $0$ that there is no correlation between the images, and a value of $1$ that the images are perfectly correlated.

The PSNR is the ratio between the maximum signal and the noise obtained from the mean squared error (MSE)~\cite{Preedanan18}. It assumes that the difference between the generated image and the reference one is noise. It is defined as follows:

\begin{equation}
\label{eq:psnr}
PSNR = 10\log_{10}(\frac{MAX^2}{MSE}),
\end{equation}

where $MAX$ is the maximum value of the image color. An accepted benchmark for the PSNR is 30 dB; a PSNR value lower than this threshold implies that the generated image presents significant deterioration~\cite{Zhao20_GAN}.

The SSIM is based on the degradation of structural information~\cite{Wang04} for the measurement of the similarity between two images~\cite{Wang11}. It is defined as follows:

\begin{equation} \label{eq:ssim}
\begin{split}
SSIM(X,Y) = (\frac{2 u_x u_y + c_1}{u_x^2 + u_y^2 + c_1 })^\alpha * (\frac{2 \sigma_x \sigma_y + c_2}{\sigma_x^2 + \sigma_y^2 + c_2 })^\beta * \\(\frac{\sigma_{xy} + c_3}{\sigma_x \sigma_y + c_3 })^\gamma,
\end{split}
\end{equation}

where $x$ and $y$ are the reference and fused images, respectively; $u_x$, $u_y$, $\sigma_x^2$, $\sigma_y^2$, and $\sigma_{xy}$ represent the mean value, variance, and covariance of images $x$ and $y$, respectively. $c_1$, $c_2$, and $c_3$ are small numbers that help to avoid a division by zero, and $\alpha$, $\beta$, and $\gamma$ are used to adjust the proportions~\cite{Zhao20_GAN}.

The possible values for the SSIM are in the range $[0,1]$, with $1$ being the best possible value.

\subsubsection{Jet Fire Segmentation}
\label{subsubsect:jet_fire_segmentation}

Following the research done by Pérez-Guerrero et al.~\cite{Perez-Guerrero21}, the Hausdorff Distance ($HD$) is used to evaluate the segmentation results obtained from the models. This metric is a dissimilarity measure that is commonly used when boundary delineation is relevant \cite{taha15}, and can be used to evaluate the resemblance between the two images when superimposed on each other \cite{huttenlocher93}. The Hausdorff Distance between two finite sections $X$ and $Y$ is defined as:

\begin{equation}
\label{eqn:haus}
HD(X,Y) = max(h(X,Y),h(Y,X)).
\end{equation}
Where $h(X,Y)$ is the directed Hausdorff distance \cite{taha15} given by:
\begin{equation}
h(X,Y) = max_{x\epsilon X}min_{y\epsilon Y}\left \| x-y \right \|.
\end{equation}

\subsubsection{Jet fire characterization}
\label{subsubsect:jet_fire_characterization}

The evaluation of the extracted geometric characteristics is done in a similar way to Pérez-Guerrero et al.~\cite{Perez-Guerrero22}, where the resulting measures are compared against the corresponding experimental data using two different error measures, the Mean Absolute Percentage Error (MAPE), and the Root Mean Square Percentage Error (RMSPE). 

The MAPE measures the percentage for the average relative differences between $N$ values from the real valued $t$ and the predicted values obtained from the segmentation $p$. It does not consider the direction of the error and is used to measure accuracy for continuous variables~\cite{Shcherbakov13}. It is defined as:

\begin{equation}
\label{eqn:mape}
MAPE = \frac{1}{N}\sum_{i=1}^{N} \frac{ |t_{i}-p_{i}|}{t_{i}} * 100.
\end{equation}

The RMSPE denotes the percentage for the relative quadratic difference between $N$ values from the real valued $t$ and the predicted values obtained from the segmentation $p$. It gives a relatively high weight to large errors~\cite{Shcherbakov13}. It is defined as:

\begin{equation}
\label{eqn:rmspe}
RMSPE = \sqrt[]{\frac{1}{N}\sum_{i=1}^{N}\left ( \frac{t_{i}-p_{i} }{p_{i}}\right )^{2}} * 100.
\end{equation}

These two metrics can also be used together to evaluate the variation of the errors. The RMSPE is usually larger or equal to the MAPE, however a greater difference between them means that there is a greater variance in the individual errors~\cite{Shcherbakov13}.

\subsection{Training and validation}
\label{subsect:train_val}

\subsubsection{Artificial IR image generation}
\label{subsubsect:training_testing_pix2pix}

For the generation of the artificial IR jet flame images, the main dataset comprised of 432 images was split into a training set of 389 images and a validation set of 43 samples. Then rotation, cropping, and mirroring operations were applied to augment the training set to a total of 6,224 training samples.

The Pix2Pix model was trained, validated and tested on an NVIDIA DGX workstation using a NVIDIA P100 GPU and the TensorFlow framework with the following hyperparamters: 100 epochs, a batch size of 4, and a learning rate of $1\times10^{-5}$ for the generator and of $1\times10^{-4}$ for the discriminator. As discussed in detail in Section~\ref{subsect:results_IR_img_GANs}, fine-tuning is not needed on these hyperparameters.

Finally, the performance and generalization capabilities of the model were further evaluated on a test set of 37 visible-IR jet flame image pairs.

\subsubsection{Jet flame segmentation and characterization}
\label{subsubsect:training_testing_seg_charact}

The PyTorch framework~\cite{pytorch} was used for the implementation of the segmentation architectures of UNet and Attention UNet. To maintain the weights as small as possible, a weight decay strategy was used with L2 regularization. The learning rate had an initial value of 0.0001 and used an ADAM optimizer during training. The loss function is a Weighted Cross-Entropy loss with weights computed according to the ENet custom class weighing scheme~\cite{paszke16}. The training was done with a batch size of 4, a learning rate of $1\times10^{-4}$ with ADAM optimizer, and up to 5000 epochs with an Early Stopping strategy to avoid over-fitting. The data was split into 80\% for training and 20\% for validation and testing. The training, validation and testing was done on an NVIDIA DGX workstation using a NVIDIA P100 GPU and further information can be found in Pérez-Guerrero et al. ~\cite{Perez-Guerrero21}.

The geometrical characteristics were extracted from the segmentation masks using the Contour Approximation Method of OpenCV \cite{opencv_library}. A pixels per metric calibration variable was established based on the distance between thermocouples in the visible images. Using the contour area, the topmost point, and the reference outlet point, the total flame length and area dimensions were calculated. The resulting measures were then compared against the corresponding experimental data using the Mean Absolute Percentage Error (MAPE), and the Root Mean Square Percentage Error (RMSPE).

\section{Experiments and discussion}
\label{sect:experiments_discussion}

\subsection{Infrared image generation using GANs}
\label{subsect:results_IR_img_GANs}

\begin{figure}[!htbp]
  \centering
  \includegraphics[scale=0.45]{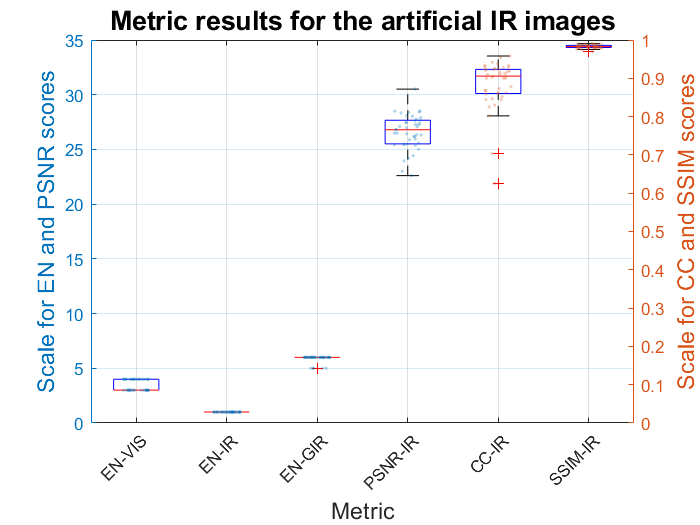}
  \caption{Results for the Pix2Pix model on the validation set.}
  \label{fig:pix2pix_val_set_results}
\end{figure}

\begin{figure*}[!htbp]
    \begin{subfigure}{0.33\textwidth}
      \includegraphics[width=\linewidth]{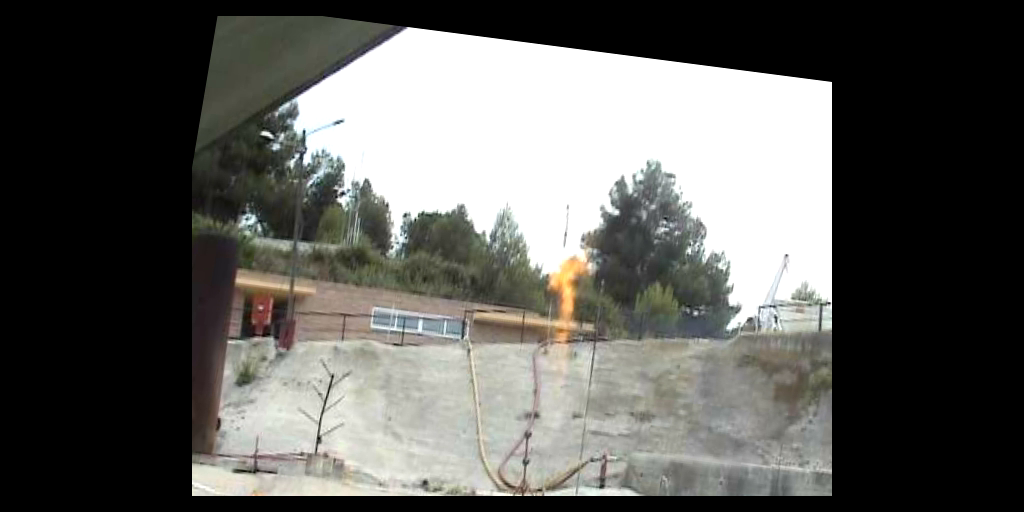}
      \caption{Visible image.}
      \label{fig:pix2pix_val_vis}
    \end{subfigure}\hfil 
    \begin{subfigure}{0.33\textwidth}
      \includegraphics[width=\linewidth]{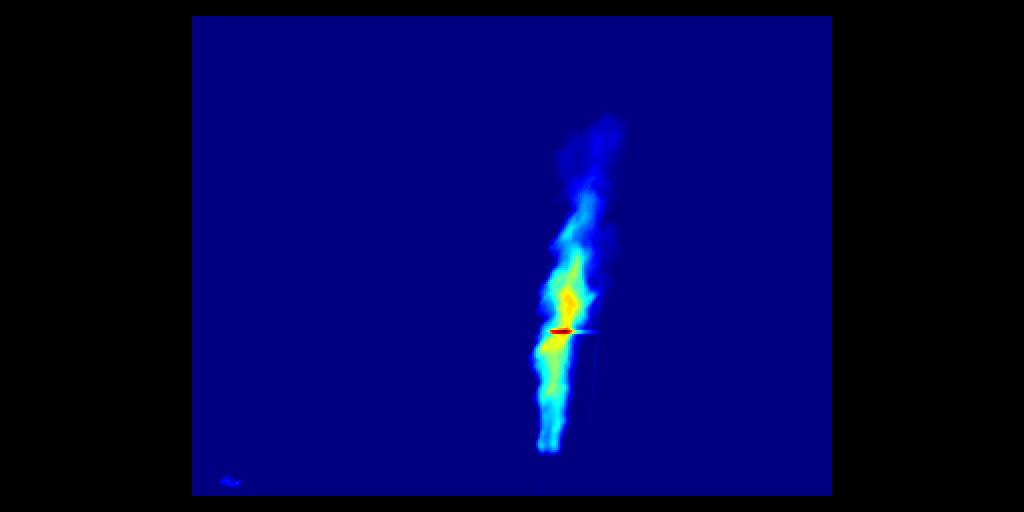}
      \caption{IR image.}
      \label{fig:pix2pix_val_ir}
    \end{subfigure}\hfil 
    \begin{subfigure}{0.33\textwidth}
      \includegraphics[width=\linewidth]{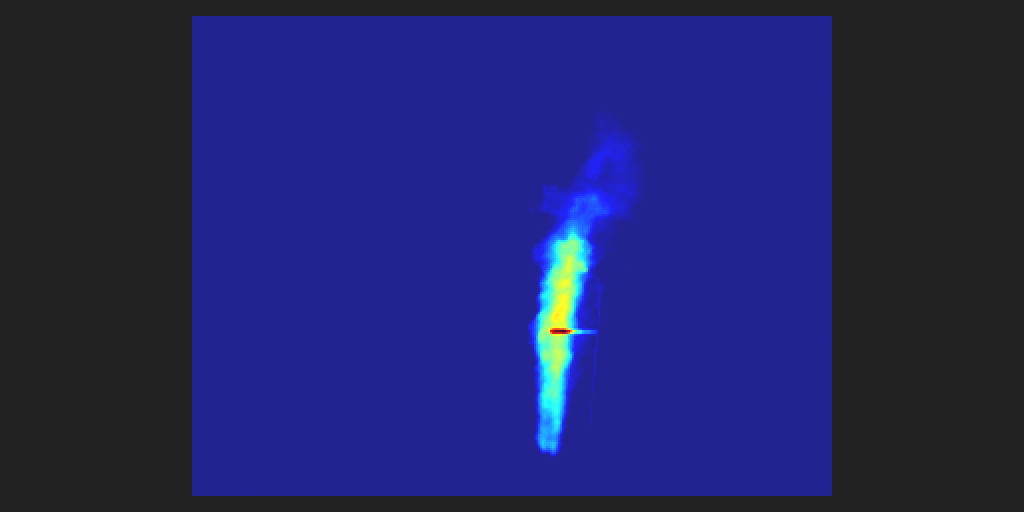}
      \caption{Generated IR image.}
      \label{fig:pix2pix_val_gir}
    \end{subfigure}\hfil 
\vspace{6pt}
    \caption{Sample resulting images of the Pix2Pix model from the validation set.}
    \label{fig:pix2pix_val_samples}
\end{figure*}

Figure~\ref{fig:pix2pix_val_set_results} shows the results for the EN, CC, PSNR, and SSIM metrics for the Pix2Pix model on the validation set. It can observed that the artificial IR images present higher values of EN, which implies more texture on these images when compared to both the visible and real IR ones. The PSNR median value is below the 30dB threshold mentioned in Section~\ref{subsect:eval_metrics}, implying that there exists a large amount of distortion on the synthetic images. However, the CC and SSIM metrics are high, with medians above the 0.9 threshold, implying a high level of linear correlation and structural similarity between the source and generated IR images. Figure~\ref{fig:pix2pix_val_samples} shows sample visible, IR, and synthetic IR images from the validation set. It is relevant to note that, given the good results obtained with the initial hyperparameters described in Section~\ref{subsubsect:training_testing_pix2pix}, it was not considered necessary to do any further fine-tuning on the model.

\begin{figure}[!htbp]
  \centering
  \includegraphics[scale=0.33]{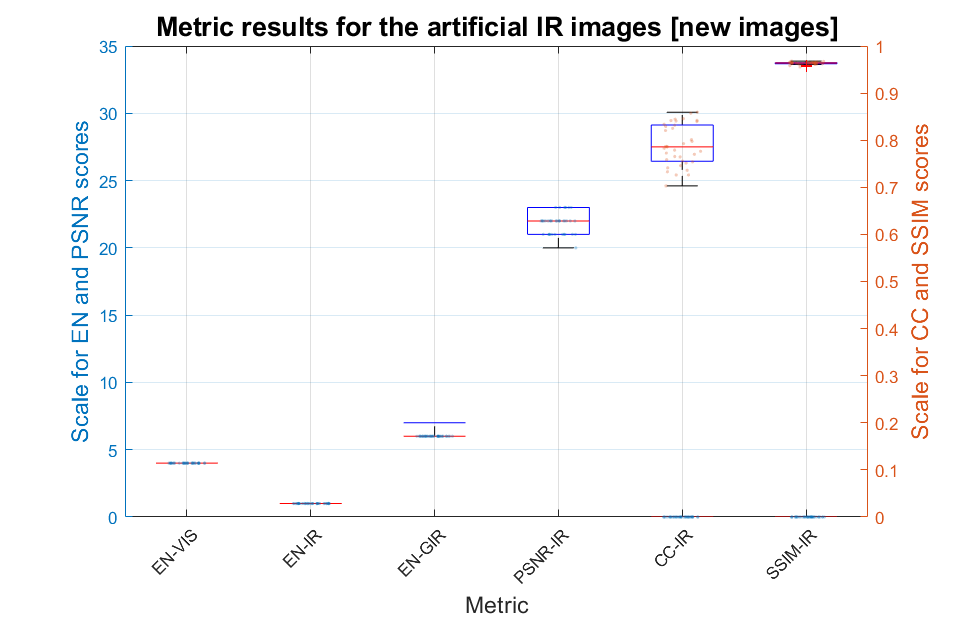}
  \caption{Results for the Pix2Pix model on the test set.}
  \label{fig:pix2pix_test_set_results}
\end{figure}

Next, Figure~\ref{fig:pix2pix_test_set_results} shows the results for the EN, CC, PSNR, and SSIM metrics for the Pix2Pix model on the test set. 

\begin{figure*}[!htbp]
    \begin{subfigure}{0.33\textwidth}
      \includegraphics[width=\linewidth]{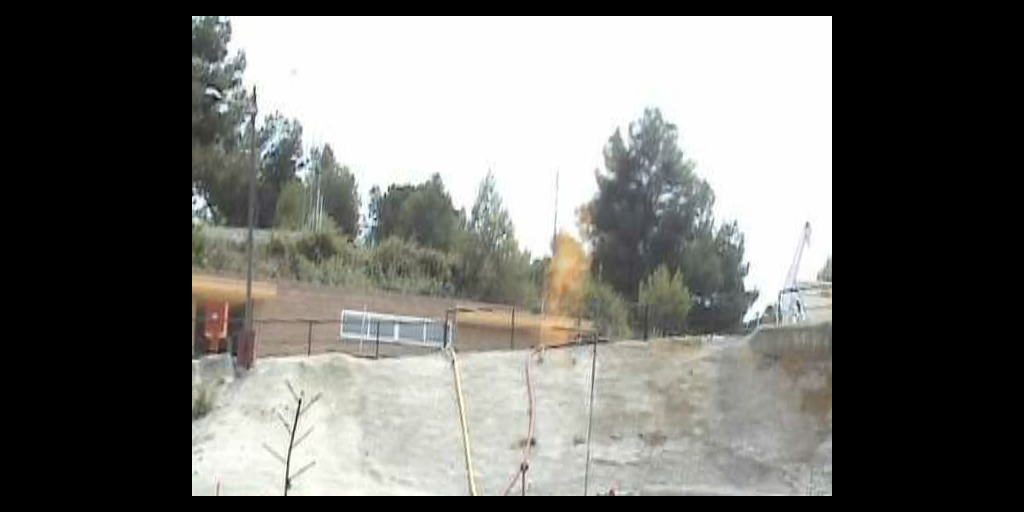}
      \caption{Visible image.}
      \label{fig:pix2pix_test_vis}
    \end{subfigure}\hfil 
    \begin{subfigure}{0.33\textwidth}
      \includegraphics[width=\linewidth]{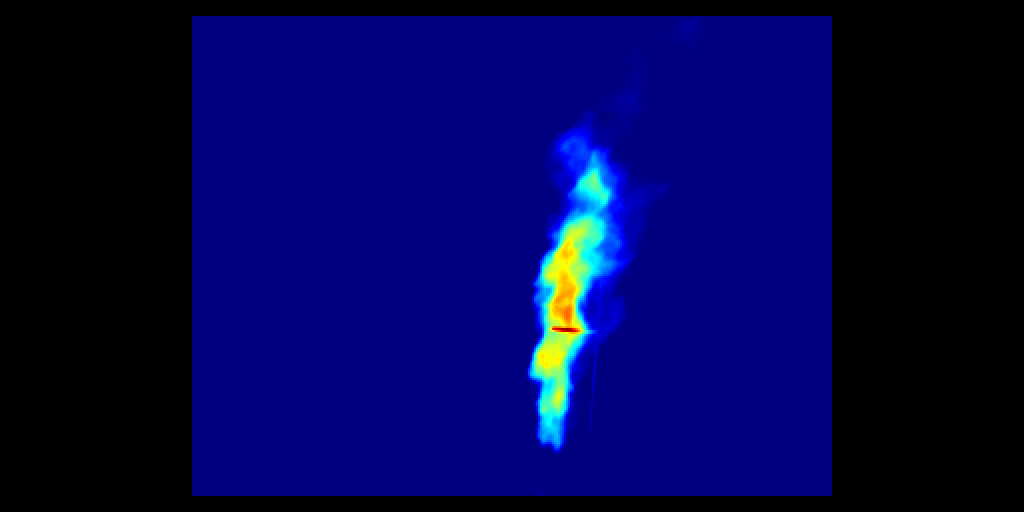}
      \caption{IR image.}
      \label{fig:pix2pix_test_ir}
    \end{subfigure}\hfil 
    \begin{subfigure}{0.33\textwidth}
      \includegraphics[width=\linewidth]{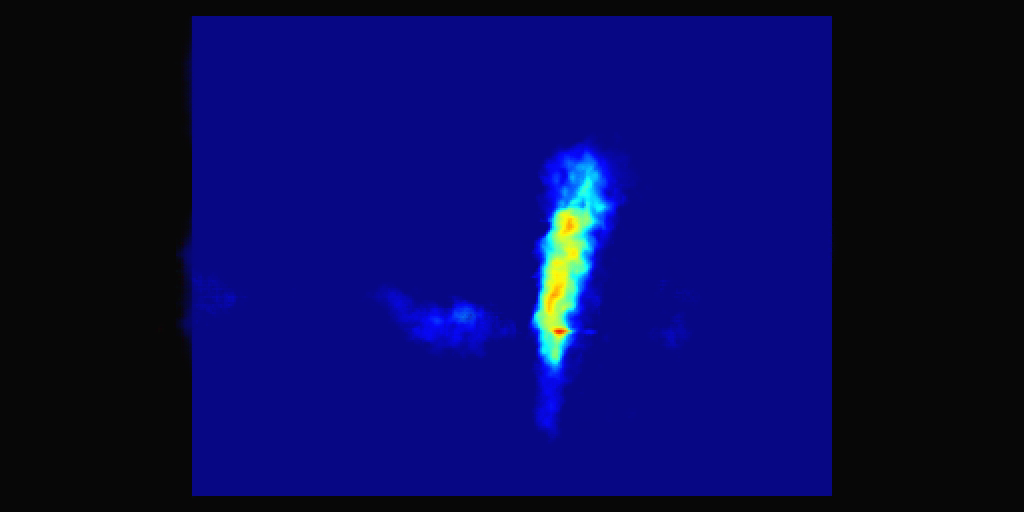}
      \caption{Generated IR image.}
      \label{fig:pix2pix_test_gir}
    \end{subfigure}\hfil 
\vspace{6pt}
    \caption{Sample resulting images of the Pix2Pix model from the test set.}
    \label{fig:pix2pix_test_samples}
\end{figure*}

As in the validation set, the artificial IR images present higher values of EN, which implies more texture on these images when compared to both the visible and real IR ones. The PSNR and CC metrics are the ones that display a higher decrease in this new test set, implying that the images have more distortion and present a lower degree of linear correlation when compared to the results of the validation set. The SSIM metric, in contrast, presents a very small decrease in performance, still being very close to a perfect score of 1. The latter implies that the real and synthetic IR images in the test set still have a significantly high level of structural similarity. Figure~\ref{fig:pix2pix_test_samples} shows sample visible, IR and synthetic IR images from the test set.

An overall decrease in the performance across all metrics can be observed, which is the expected behavior on the new images of the test set. However, the mentioned decrease in performance is small, which speaks well of the generalization capabilities of the model for jet flame imagery.

\subsection{Image segmentation and jet flame characterization}
\label{subsect:results_img_seg_charact}
\begin{figure}[!htbp]
  \centering
  \begin{subfigure}{0.23\textwidth}
      \includegraphics[width=\linewidth]{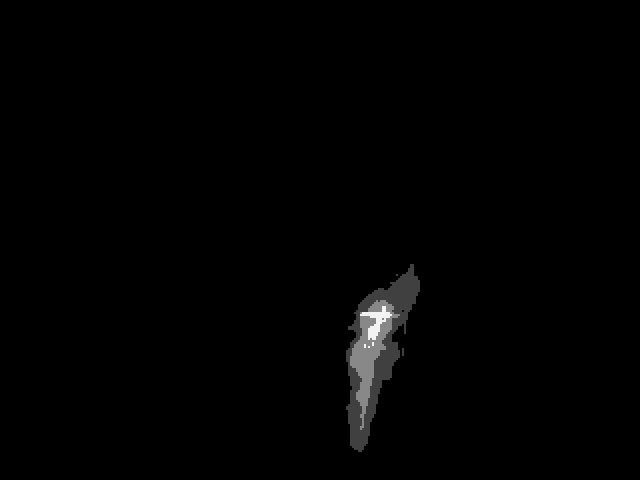}
      \caption{UNet segmentation with brightness adjustment methods.}
    \end{subfigure}
    \begin{subfigure}{0.23\textwidth}
      \includegraphics[width=\linewidth]{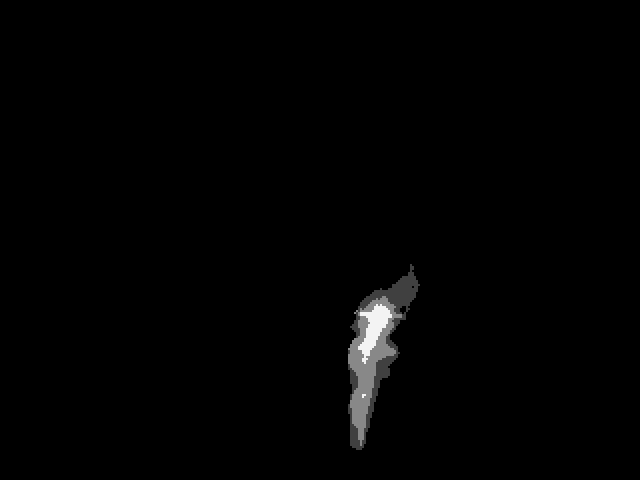}
      \caption{Attention UNet segmentation with brightness adjustment methods.}
    \end{subfigure}
    \begin{subfigure}{0.23\textwidth}
      \includegraphics[width=\linewidth]{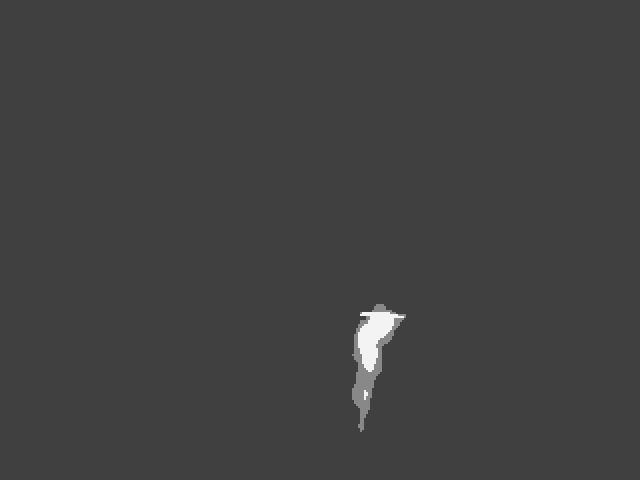}
      \caption{UNet segmentation without brightness adjustment methods.}
    \end{subfigure}
    \begin{subfigure}{0.23\textwidth}
      \includegraphics[width=\linewidth]{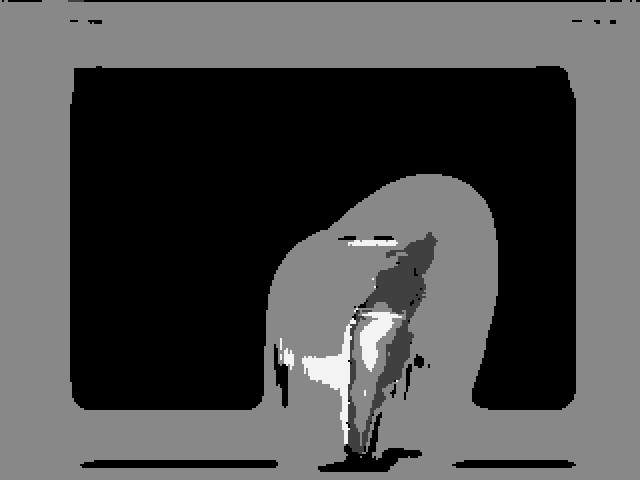}
      \caption{Attention UNet segmentation without brightness adjustment methods.}
    \end{subfigure}
  \caption{Difference in segmentation with or without brightness adjustment methods using the UNet and Attention UNet models.}
  \label{fig:brightness_diff}
\end{figure}

The initial segmentation results without brightness adjustment methods caused severe artifacts when using generated IR images that showed high values of brightness. A sample can be observed in Figure \ref{fig:brightness_diff}. The UNet model was more sensitive to the differences in brightness than the Attention UNet model, which produced less instances of segmentation masks with artifacts.

Additionally, the segmentation obtained with Attention UNet produced a lower mean Hausdorff distance than the segmentation obtained with UNet. The Hausdorff distance was calculated between the segmentation masks produced from the real and generated IR images. The overall results are summarized in Table \ref{tab:hausdorff}.

\begin{table}[!htbp]
\centering
\caption{The mean Hausdorff distance between the segmentation obtained from the original IR image and the one obtained from the generated IR image, for each one of the segmentation models.}
\label{tab:hausdorff}
\begin{tabular}{c|cc|cc}
\hline
\textbf{Experiment}                                          & \multicolumn{2}{c|}{\textbf{JFP 05 11}}                                           & \multicolumn{2}{c}{\textbf{JFP 05 12}}                                            \\ \hline
\textbf{Model}                                               & \textbf{UNet} & \textbf{\begin{tabular}[c]{@{}c@{}}Attention\\ UNet\end{tabular}} & \textbf{UNet} & \textbf{\begin{tabular}[c]{@{}c@{}}Attention\\ UNet\end{tabular}} \\ \hline
\begin{tabular}[c]{@{}c@{}}Hausdorff\\ distance\end{tabular} & 455.919       & 404.223                                                           & 755.506       & 685.094                                                           \\ \hline
\end{tabular}
\end{table}

\subsubsection{Flame total height}

\begin{figure}[!htbp]
  \centering
  \begin{subfigure}{0.35\textwidth}
      \includegraphics[width=\linewidth]{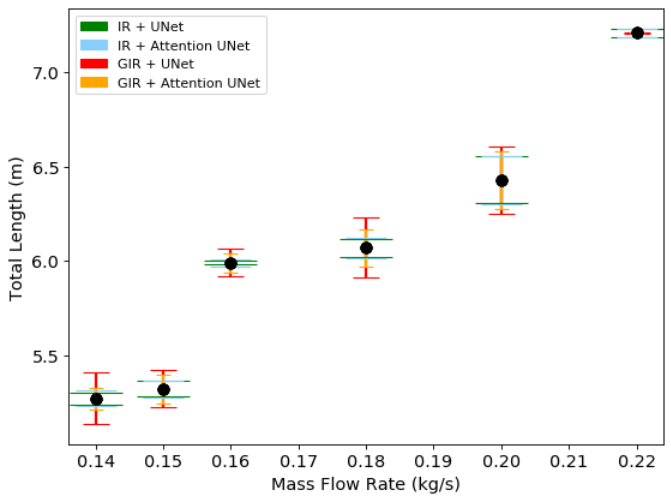}
      \caption{Results for JFP 05 11}
    \end{subfigure}
    \begin{subfigure}{0.35\textwidth}
      \includegraphics[width=\linewidth]{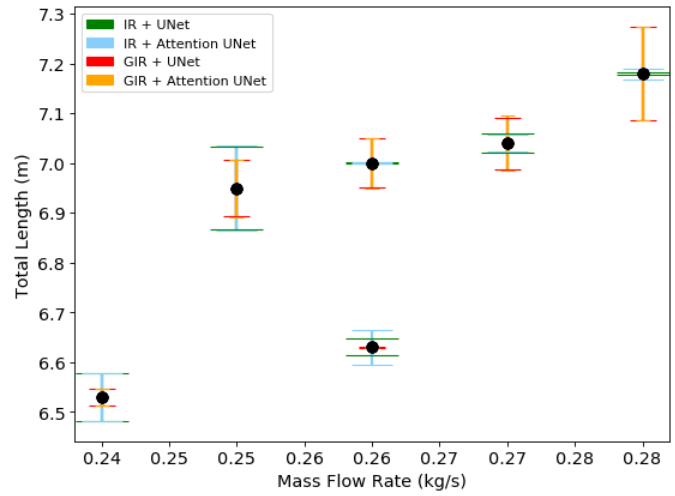}
      \caption{Results for JFP 05 12}
    \end{subfigure}
  \caption{Total flame length percentage change between the experimental values and the results obtained from the segmentation of real IR and generated IR (GIR) images using the UNet and Attention UNet models.}
  \label{fig:errorbars_len}
\end{figure}

Table \ref{tab:length} and Figure~\ref{fig:errorbars_len} show the results for the total flame length obtained from the segmentation masks generated by the UNet and Attention UNet models. The segmentation was done for both the original and generated IR images. It can be observed that the generated IR images tend to have higher error values than the original IR images; however the difference is not too large and the highest MAPE of 10.741\% still allows for a decent representation of the total flame length. These results are also congruent with the experiments shown in Pérez-Guerrero et al. ~\cite{Perez-Guerrero22}, where the highest MAPE for height had a value of 11.1\%.

UNet, in general, obtained the lowest MAPE when extracting the total flame length from the original IR images. However, Attention UNet showed the smallest difference between the total flame length obtained from the original and generated IR images, which can be due to the sensitivity of UNet to slight brightness differences of the IR images as previously discussed.

\begin{table}[!htbp]
\centering
\caption{The error of the flame total height obtained from different segmentation models and for each experiment analysed.}
\label{tab:length}
\begin{tabular}{ccccc}
\hline
\multicolumn{5}{c}{\textbf{Total Flame Length}}                                                                                                       \\ \hline
\multicolumn{1}{l|}{\textbf{Experiment}} & \multicolumn{2}{c|}{\textbf{JFP 05 11}}                     & \multicolumn{2}{c}{\textbf{JFP 05 12}} \\ \hline
\multicolumn{1}{c|}{\textbf{IR Image}}   & \textbf{Original} & \multicolumn{1}{c|}{\textbf{Generated}} & \textbf{Original} & \textbf{Generated} \\ \hline
\multicolumn{5}{c}{\textbf{UNet}}                                                                                                               \\ \hline
\multicolumn{1}{c|}{MAPE}                & 4.591\%           & \multicolumn{1}{c|}{10.741\%}           & 2.855\%           & 4.494\%            \\ \hline
\multicolumn{1}{c|}{RMSPE}               & 5.952\%           & \multicolumn{1}{c|}{12.258\%}           & 4.076\%           & 5.383\%            \\ \hline
\multicolumn{5}{c}{\textbf{Attention UNet}}                                                                                                     \\ \hline
\multicolumn{1}{c|}{MAPE}                & 5.095\%           & \multicolumn{1}{c|}{7.323\%}            & 3.292\%           & 4.601\%            \\ \hline
\multicolumn{1}{c|}{RMSPE}               & 6.228\%           & \multicolumn{1}{c|}{8.660\%}            & 4.327\%           & 5.498\%            \\ \hline
\end{tabular}
\end{table}

\subsubsection{Flame area}

\begin{figure}[!htbp]
  \centering
  \begin{subfigure}{0.35\textwidth}
      \includegraphics[width=\linewidth]{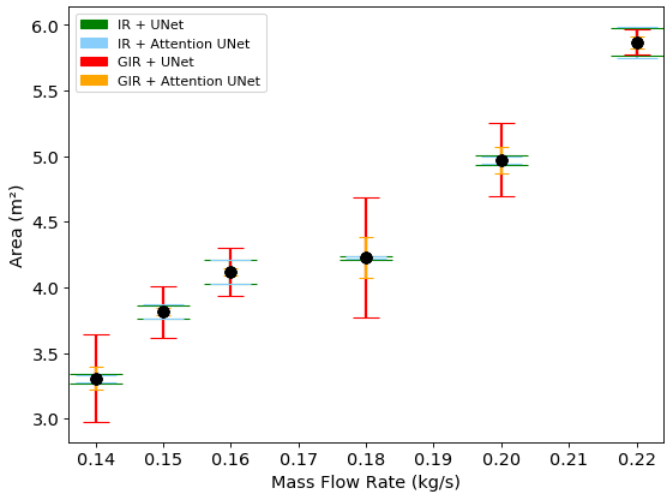}
      \caption{Results for JFP 05 11}
    \end{subfigure}
    \begin{subfigure}{0.35\textwidth}
      \includegraphics[width=\linewidth]{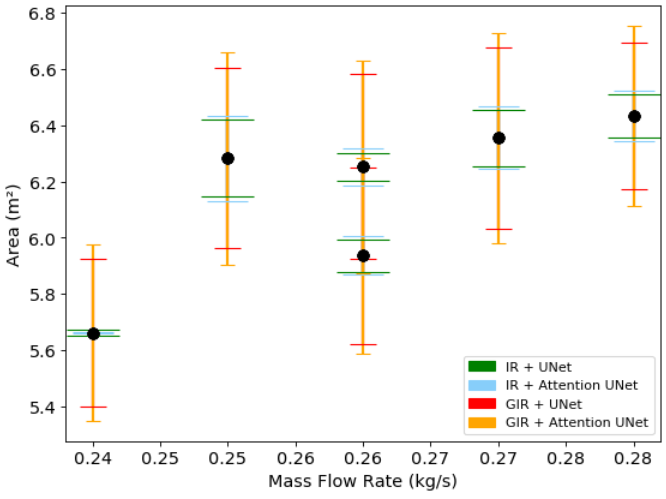}
      \caption{Results for JFP 05 12}
    \end{subfigure}
  \caption{Flame area percentage change between the experimental values and the results obtained from the segmentation of real IR and generated IR (GIR) images using the UNet and Attention UNet models.}
  \label{fig:errorbars_area}
\end{figure}

Table \ref{tab:area} and Figure~\ref{fig:errorbars_area} show the results for the area obtained from the segmentation masks generated by the UNet and Attention UNet models. The segmentation was done for both the original and generated IR images. In a similar way to the total length, it can be observed that the generated IR images tend to have higher error values than the original IR images, with the highest MAPE value being 35.208\%. These results, however, are somewhat congruent with the experiments shown in Pérez-Guerrero et al. ~\cite{Perez-Guerrero22}, where the highest MAPE for area was a value of 21.9\%. 

UNet once again obtained the lowest MAPE when extracting the area from the original IR images, however, the difference from the results of Attention UNet is minimal. The only case where the area obtained from the generated IR images did not go beyond a MAPE of 10\%  was with the "JFP 05 11" experiment and using the Attention UNet model, obtaining a value of 7.435\%. This may be because the flames in the "JFP 05 12" experiment were not as clearly visible as the ones in the "JFP 05 11" experiment so certain parts of the fire were not fully represented in the generated IR image, resulting in high error values for the area.

\begin{table}[!htbp]
\centering
\caption{The error of the flame area obtained from different segmentation models and for each experiment analysed.}
\label{tab:area}
\begin{tabular}{ccccc}
\hline
\multicolumn{5}{c}{\textbf{Flame Area}}                                                                                                               \\ \hline
\multicolumn{1}{l|}{\textbf{Experiment}} & \multicolumn{2}{c|}{\textbf{JFP 05 11}}                     & \multicolumn{2}{c}{\textbf{JFP 05 12}} \\ \hline
\multicolumn{1}{c|}{\textbf{IR Image}}   & \textbf{Original} & \multicolumn{1}{c|}{\textbf{Generated}} & \textbf{Original} & \textbf{Generated} \\ \hline
\multicolumn{5}{c}{\textbf{UNet}}                                                                                                               \\ \hline
\multicolumn{1}{c|}{MAPE}                & 5.502\%           & \multicolumn{1}{c|}{25.845\%}           & 7.190\%           & 30.119\%           \\ \hline
\multicolumn{1}{c|}{RMSPE}               & 6.371\%           & \multicolumn{1}{c|}{28.366\%}           & 8.203\%           & 30.254\%           \\ \hline
\multicolumn{5}{c}{\textbf{Attention UNet}}                                                                                                     \\ \hline
\multicolumn{1}{c|}{MAPE}                & 5.470\%           & \multicolumn{1}{c|}{7.435\%}            & 8.158\%           & 35.208\%           \\ \hline
\multicolumn{1}{c|}{RMSPE}               & 6.783\%           & \multicolumn{1}{c|}{8.734\%}            & 9.344\%           & 35.311\%           \\ \hline
\end{tabular}
\end{table}

\subsection{Discussion and future work}
\label{subsect:discussion_future_work}
The characterization of large-scale jet flames through computer vision is a promising approach following recent strides in Deep Learning-based analysis in this regard. However, these models are heavily dependent on the amount of data and the quality of the labels available for training, validation, and testing. The data acquisition of jet fires is done at a high cost since it involves expensive experiments. The inclusion of IR imagery has a positive impact on the thorough analysis of the jet fires' characteristics; however, this only adds another layer of complexity and costliness to the experiments due to expensive IR cameras and the necessity of coupling both visible and infrared images with cameras that usually have different resolutions and frame rates.

Thus, the generation of artificial IR images from visible ones can significantly reduce the complexity and cost of the data acquisition of jet fire images for DL-based characterization. In recent times, GANs have displayed state-of-the-art performance for different data augmentation and synthetic image generation applications.

In the present paper, the feasibility of using GANs to generate artificial IR images was demonstrated, along with their capability to be integrated into the data acquisition stage of a DL-based jet fire characterization process. A pipeline for the generation of synthetic IR images from visible ones was presented, followed by the semantic segmentation of the radiation zone of the flames and, finally, the characterization of their geometric characteristics.

Using the Pix2Pix model for visible-to-IR jet fire image translation, a decent approximation of the thermal information provided by real infrared imagery can be obtained. Said results, coupled with an Attention UNet network, allow us to attain very similar segmentation and characterization results compared to experiments with real IR imagery for the characterization of the total length of the flames.

Nevertheless, the explored GAN-based approach does not perform adequately for the characterization of the total area of the flames. This limitation is potentially due to the visibility of the jet flame in the visible images; the model is unable to synthesize appropriate IR images when the fire is not clearly visible (e.g., due to environment lights or flame occlusion). This problem could potentially be overcome in future works by the inclusion of more diverse and challenging images on the training set for the GAN-based image generation or through the implementation of few-shot learning strategies.

\section{Conclusions}
\label{sect:conclusions}

 The Pix2Pix model occasionally generated synthetic IR images that displayed a considerably higher brightness level when compared to the real ones. This difference in brightness created artifacts in the segmentation of the flames, which the UNet model being more sensitive to this problem than the Attention UNet model. An analysis of the RMS brightness level between the real and artificial IR images is recommended to make corrections accordingly.

The results for the total flame length are congruent with the previous work and represent a decent approximation to the experimental data. Attention UNet obtained the smallest difference between the results extracted from the original and generated IR images. This means that the proposed pipeline could be applied with confidence for the characterization of the flame's total height, employing only visual light imagery. 

The results for the total flame area, were also somewhat congruent with the results of the previous work, even when showing a larger difference between the values obtained from the original and generated IR images. Overall, the Attention UNet model was observed to obtain more consistent segmentation results than the UNet model, therefore it would be advised to use the Attention UNet model in combination with the generated IR images. The results look promising and the models could be employed with confidence in a smart sensing pipeline with visible cameras.

\section*{Acknowledgments}
The authors wish to thank the AI Hub and the Centro de Innovación de Internet de las Cosas at Tecnológico de Monterrey for their support for carrying the training and testing of the computer models explored in this paper with their NVIDIA DGX workstation.

The authors also wish to thank the Spanish Ministry of Economy and Competitiveness, and the Institut d’Estudis Catalans for providing the jet fire data used in the experiments presented in this paper.

This work was supported in part by the SEP-CONACYT ECOS-ANUIES Project 315597.

\bibliographystyle{cas-model2-names}

\bibliography{references}

\end{document}